\documentclass[letterpaper]{article} 
\usepackage{aaai2026}
\usepackage{times}  
\usepackage{helvet}  
\usepackage{courier}  
\usepackage[hyphens]{url}  
\usepackage{graphicx} 
\usepackage{natbib}  
\usepackage{caption} 
\usepackage{algorithm}
\usepackage{algorithmic}

\usepackage{booktabs} 

\usepackage{tabularx}  

\usepackage{xcolor}

\definecolor{AIOColor} {RGB}{34, 125, 34}

\definecolor{FSColor}{RGB}{102, 0, 204}

\newcommand{\answerYes}[1]{{#1}} 
\newcommand{\answerNo}[1]{\textcolor{teal}{#1}} 
\newcommand{\answerNA}[1]{\textcolor{gray}{#1}}

\usepackage{tcolorbox}
\usepackage{enumitem}

\usepackage{newfloat}
\usepackage{listings}
\DeclareCaptionStyle{ruled}{labelfont=normalfont,labelsep=colon,strut=off} 
\floatstyle{ruled}
\newfloat{listing}{tb}{lst}{}
\floatname{listing}{Listing}
\usepackage[draft,inline,nomargin]{fixme}
\fxsetup{theme=colorsig,mode=multiuser,inlineface=\itshape,envface=\itshape}
\FXRegisterAuthor{jg}{ajg}{Jeffrey}

\title{Auditing Google's AI Overviews and Featured Snippets:\\A Case Study on Baby Care and Pregnancy}

\author{
    Desheng Hu\textsuperscript{\rm 1},
    Joachim Baumann\textsuperscript{\rm 2},
    Aleksandra Urman\textsuperscript{\rm 1},
    Elsa Lichtenegger\textsuperscript{\rm 1},
    Robin Forsberg\textsuperscript{\rm 1,3},
    Aniko Hannak\textsuperscript{\rm 1,4},
    Christo Wilson\textsuperscript{\rm 5}
}

\affiliations{
    \textsuperscript{\rm 1}University of Zurich, Zurich, Switzerland\\
    \textsuperscript{\rm 2}Stanford University, Stanford CA, USA\\
    \textsuperscript{\rm 3}University of Helsinki, Helsinki, Finland\\
    \textsuperscript{\rm 4}Corvinus University of Budapest, Budapest, Hungary\\
     \textsuperscript{\rm 5}Northeastern University, Boston MA, USA\\
    desheng@ifi.uzh.ch, joachimbaumann@stanford.edu, urman@ifi.uzh.ch, lichtenegger@ifi.uzh.ch,\\
    robinchristopher.forsberg@uzh.ch, hannak@ifi.uzh.ch, cbw@ccs.neu.edu
}

\usepackage{bibentry}

\begin{document}

\maketitle

\begin{abstract}

Google Search increasingly surfaces AI-generated content through features like AI Overviews (AIO) and Featured Snippets (FS), which users frequently rely on despite having no control over their presentation. Through a systematic algorithm audit of 1,508 real baby care and pregnancy-related queries, we evaluate the quality and consistency of these information displays. Our robust evaluation framework assesses multiple quality dimensions, including answer consistency, relevance, presence of medical safeguards, source categories, and sentiment alignment. Our results reveal concerning gaps in information consistency, with information in AIO and FS displayed on the same search result page being inconsistent with each other in 33\% of cases. Despite high relevance scores, both features critically lack medical safeguards (present in just 11\% of AIO and 7\% of FS responses). While health and wellness websites dominate the source categories for both AIO and FS, FS more often link to commercial sources. These findings have important implications for public health information access and demonstrate the need for stronger quality controls in AI-mediated health information. Our methodology provides a transferable framework for auditing AI systems across high-stakes domains where information quality directly impacts user well-being.

\end{abstract}

\begin{figure*}[t!] 
\centering
  \includegraphics[width=0.75\textwidth]{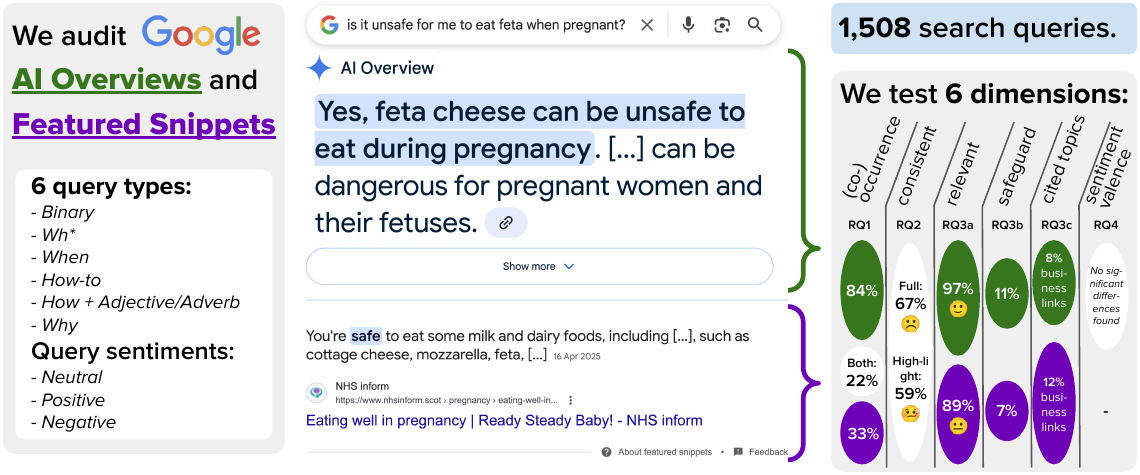}
  \caption{Overview of our audit study {methods and results} on Google's {\color{AIOColor}\textbf{AI Overviews (AIO)}} and {\color{FSColor}\textbf{Featured Snippets (FS)}} for pregnancy and baby care information. {In the example screenshot, the AIO appears above the FS. The first sentence of the AIO is highlighted by Google to add emphasis.} 
{We find that AIO occur more frequently than FS (\textbf{RQ1}) and show considerable inconsistency with FS answers, more profound in highlighted pairs (\textbf{RQ2}). While AIO and FS responses are generally relevant (\textbf{RQ3a}), they provide safeguard cues infrequently (\textbf{RQ3b}), and FS sources concern significantly higher proportions from commercial categories than AIO or ten blue links results (\textbf{RQ3c}).
 \textbf{RQ4:} We do not find evidence of ``confirmation bias'' in AIO answers, where a user's sentiment (positive/neutral/negative) is reflected in the query.}
  }
  \label{fig:figure1}
  \centering
\end{figure*}

\section{Introduction}

Search engines like Google serve as critical information gatekeepers for high-stakes topics where people need to rapidly acquire extensive knowledge to make important decisions. This includes domains like healthcare, finance, and politics, where search results significantly impact individual and societal well-being~\cite{fox2011health,fernandez2024search}. Among these areas, baby care and pregnancy offer a compelling example -- new and expectant parents make frequent, time-sensitive decisions while learning from many new concepts, often relying on search engines during a critical period lasting months or years. Specifically, 85\% of pregnant women {surveyed by \cite{giacometti_internet_2024}} turned to Google for pregnancy-related information in 2024. While over 80\% stated that the information they found online had a medium to very high influence on their decision-making related to pregnancy, only 42\% reported validating such information with their doctors \cite{giacometti_internet_2024}. Thus, web search, particularly Google, has become a primary source for crucial pregnancy and baby care information.
Many users rely on these search results as standalone resources without seeking  medical professionals' validation.
The potential risks to expectant parents and newborns from false or low-quality information make it essential to scrutinize the quality of search outputs on these topics.

Prior work on the quality of web search results related to baby care and pregnancy is very limited and focused on few particular aspects of baby care/pregnancy such as pregnancy-related nausea, examining information quality in the ``ten blue links" -- websites linked on Google Search Engine Result Pages (SERPs) \cite{monje_analysis_2023,sacks_how_2013,artieta-pinedo_evaluation_2018}. However, in recent years, Google Search has evolved with many new components added to SERPs~\cite{oliveira2023evolution}, including featured snippets (FS) and AI Overviews (AIO) -- visual boxes containing direct answers powered by information retrieval and generative LLMs, respectively (Figure 1).

{AIO synthesizes information from multiple webpages into a short summary that sits above organic results and includes links back to source pages, which Google describes as a “jumping-off point” to help users grasp complex queries. Released in the U.S. in May 2024, by mid-2025 AIO reached over 200 countries with 2+ billion monthly users~\cite{google_dev_aifeatures,techcrunch2025aio}. AIO has been criticized for occasionally producing misleading or harmful information \cite{Williams2024whyAIOwrong} which, coupled with AIO's rapid deployment, underscores the urgency of assessing the quality and safety of such direct-answer features.}

Research on information quality within these elements, however, is scarce, both for pregnancy/baby care and more generally. {Despite Google's YMYL (Your Money or Your Life) policy requiring enhanced scrutiny for health-related content~\cite{google2023sqrg}, top-ranked results can still contain false health claims.
Prior research shows this contributes to misinformation spread \cite{cai2021using}. Even 
websites rated as high-quality by work \cite{lin2023high}, like Healthline, now include AI-generated articles that may contain unreliable medical information \cite{dupre2024wikipedia}.}

{Another research gap concerns how query characteristics like question type and query sentiment relate to the display and quality of AIO/FS answers.} This gap is significant given  prior findings on query formulation effects. Question-based queries help users achieve their goals with minimal reformulation~\cite{vanderschantz2017study} and trigger better direct answers on Google~\cite{strzelecki2020direct}. Question rewriting improves QA system performance~\cite{buck2017ask,chu2020ask}. Certain question types (e.g., where-questions) yield higher-quality answers than others (e.g., how-questions)\cite{zhao2019evaluation}. Query sentiment presents additional concerns. Research shows that sentiment in queries can influence responses in Generative AI search engines, with responses aligning with query bias~\cite{venkit2024search}. {Such bias is particularly dangerous in high-stakes domains~\citep{li2024generative,baumann2025reduced}.
Consider pregnancy-related queries like ``is
it safe/unsafe for pregnant women to take [medicine/food].'' Answers that mistake ``safe'' for “unsafe” may cause severe health consequences. Therefore, high-stakes health information should not depend on query sentiment polarity.
These concerns, combined with the above evidence, }highlight the critical need to examine how question type and sentiment influence both the appearance and quality of AI-driven components like AIO and FS in high-stakes domains such as baby care and pregnancy.

\paragraph{Research Questions.}
These identified research gaps directly inform our study's focus and specific RQs outlined below. The limited research on modern SERP components such as AI Overviews and featured snippets in high-stakes domains (addressed in RQ1 and RQ3) represents a critical oversight, especially given their increasing prominence in search results. Similarly, the lack of comparative analysis when these components co-occur (addressed in RQ2) limits our understanding of potential information inconsistencies that could affect user decision-making. Finally, the unexplored relationship between query sentiment and response characteristics (addressed in RQ4) may reveal concerning biases in how information is delivered to users. By systematically investigating these questions in the context of pregnancy and baby care searches, we aim to provide insights with implications beyond this specific domain to information quality in high-stakes search contexts more broadly. {While we focus on one domain rather than multiple topics, this approach enables deeper analysis with context-specific metric design and extensive manual annotation. It also provides an evaluation framework for AI-generated search outputs that can be transferred to other domains.} Our specific RQs are formulated as follows:

\begin{itemize}[leftmargin=3mm]
\item  \textbf{RQ1:} How prevalent are AIO and FS overall and across different question types and query sentiments in baby care and pregnancy searches?
\item  \textbf{RQ2:} When AIO and FS co-occur on the same SERP, how consistent is the information they provide overall and across different question types and sentiments?
\item  \textbf{RQ3:} What is the response quality of FS and AIO in terms of  output relevance (RQ3a), presence of safeguards (RQ3b), and source categories (RQ3c) overall and across different question types and sentiments?~\looseness=-1
\item \textbf{RQ4:} How is query sentiment associated with the sentiment of responses from AIO and FS when query topic and question type are controlled?

 \end{itemize}

To address our research questions, we built a pipeline involving query collection, reformulation, and algorithm auditing experiments to test how query factors influence AI Overviews (AIO) and featured snippets, while controlling for variables such as location, browser, and settings. We also developed evaluation metrics and a manual annotation codebook to assess result quality. The dataset, pipeline and metric annotation codebook will be made publicly available upon publication to contribute to further research in this field.

We find  that in the context of baby care and pregnancy, AIOs appear much more frequently than FS (84\% vs. 33\%) overall, with both co-occurring in 22\% of cases. There is a significant association between the frequency of AIO and FS appearance on SERP and query characteristics (question type and query sentiment) (\textbf{RQ1}). When AIO and FS co-occur on the same SERP, the information provided in them is relatively often inconsistent (33\% overall, 41\% for highlighted content), with consistency varying significantly by question type but not by sentiment (\textbf{RQ2}). Both AIO and FS are relevant to the search query in the  majority of the cases (97\% and 89\%, respectively), and relevance is significantly associated with question type for FS only (\textbf{RQ3a}). Safeguard cues appear infrequently (11\% in AIO, 7\% in FS) , with  question type being significantly associated with safeguard presence in AIO only (\textbf{RQ3b}). Both AIO and FS cite more health-related sources than regular ten blue links, but FS draws disproportionately from business and shopping sites; {manual credibility assessment of top 10\% domains reveals low- and medium-credibility sources account for nearly half of AIO/FS citations}; source categories are associated with question type but not sentiment (\textbf{RQ3c}). Meanwhile, we find no evidence of confirmation bias in AIO responses when controlling for query topic and question type (\textbf{RQ4}). Finally, results based on these metrics reveal important implications, and the metrics are transferable to audits of other high-stakes domains such as legal or political information seeking.

\section{Related Work}

\subsection{Featured Snippet Measurement/Audit}

Featured snippets, which are pieces of information extracted from webpages and highlighted at the top of search results (prior to Google's introduction of AIO), aim to directly address users' information-seeking needs \cite{bink2022featured}. They are typically paired with highlighted text from the source page. Prior work shows that the most common source domain for generating featured snippets is Wikipedia, and they are often triggered by question queries \cite{strzelecki2019featured,strzelecki2020direct}.

On one hand, this feature has been shown to attract more attention and increase users’ dwell time on the search engine result page (SERP) \cite{wu2020providing,gleason2023google}, triggering ``good abandonment'' -- defined as when users’ information needs are satisfied without clicking on a SERP result \cite{li2009good}. On the other hand, their presence can lead users to overestimate the trustworthiness of the information \cite{bink2022featured}, and may contribute to health anxiety in health-related searches, particularly when the displayed results are distressing \cite{bink2022featured}. This is especially concerning when featured snippets may spread misleading information in domains such as health or political information seeking \cite{lurie2021searching,scull2020dr}.

Among these studies, \citet{strzelecki2019featured,strzelecki2020direct,scull2020dr,zhao2019evaluation} are the most relevant to our work. It's been shown that the appearance of featured snippets is closely related to the question form of queries \cite{strzelecki2019featured,strzelecki2020direct}, and that the question type of the query is associated with answer quality (when measured via a singular quality score) \cite{zhao2019evaluation}. Meanwhile, Wikipedia has been found to be the most sourced domain over a large query set (implying that the information previewed in FS is very likely to be correct)~\cite{strzelecki2019featured,strzelecki2020direct}. Featured snippet source domains for health-related queries are also mostly reliable websites, such as Mayo Clinic and familydoctor.com~\cite{scull2020dr} These studies motivate our RQ1 and RQ3 regarding investigation of question type's association with FS's appearance and answer quality when measured across several dimensions (including source domains).

\subsection{Generative AI search engine audit}

Studies specifically auditing and evaluating generative AI search engines are limited, with few early explorations conducting evaluation via either fully manual annotation~\cite{li2025human,liu2023evaluating,hu2024evaluating} or hybrid methods combining manual effort with LLM-as-a-judge or machine learning classifiers~\cite{venkit2024search,li2024generative}. Most focus on specific answer quality properties (e.g., verifiability, sentiment, source category) or scenarios (e.g., adversarial attacks). For example, prior work~\cite{liu2023evaluating} examined verifiability through broad metrics (fluency, perceived utility) for measuring answers' text generally and specific metrics (citation recall, citation precision) for checking answers' reference, finding that generative AI responses often exhibit high fluency and perceived utility but frequently contain unsupported statements or inaccurate citations. Another study~\cite{hu2024evaluating} evaluated generative search engines under adversarial input scenarios, demonstrating effectiveness in inducing incorrect responses via manipulative questions. Research in~\citet{li2024generative} examined sentiment in queries and answers, as well as source types, finding evidence of response sentiment bias driven by query sentiment, plus commercial and geographic biases in source categories. Research in~\citet{li2025human} specifically studied people's trust in GenAI search engines compared to traditional ones via a preregistered, randomized experiment on a large U.S. representative sample.

Among these works,~\citet{venkit2024search} conducted a more comprehensive evaluation spanning answer text, citations, sources, and user interface via hybrid methods including LLM-as-a-judge and manual evaluation (user interviews). They found that GenAI answers tend to use overtly confident language and align with biases implied in questions (``confirmation bias"). However, while confirmation bias wasn't specifically studied during implementation in~\cite{venkit2024search}, and~\cite{li2024generative} did not control for specific topics or question types -- factors known to affect question-answering systems~\cite{li2025human,zhao2019evaluation,buck2017ask,chu2020ask} -- we investigate confirmation bias more rigorously by controlling query topic and question type (RQ4). We also integrate metrics from prior studies relevant to our research setting, which we elaborate on in the following methodology section. Previous work has raised concerns about biases in LLM-as-a-judge evaluation~\cite{ye2024justice} and documented only moderate correlation between human annotation and LLM-as-a-judge (e.g., Pearson correlation of 0.62 in similar search settings~\cite{venkit2024search}), while recent evidence has documented systematic risks and unreliability when using LLMs for annotation tasks without rigorous validation~\cite{baumann2025llmhacking}. Given these concerns, we opt for full manual evaluation for all metrics except source category in this study.

\section{Methodology}

To address our research questions, we built a pipeline involving query selection, reformulation, and algorithm auditing experiments to test how query factors influence AI Overviews (AIO) and featured snippets, while controlling for variables such as location, browser, and settings. For RQ1 (prevalence of AI Overviews and Featured Snippets), we measured component appearance rates across search results collected using our auditing pipeline for the diverse query set we constructed. For RQs 2, 3, 4, we evaluated the AIO/FS content relying on manual annotation and the specialized metrics within the codebook we developed. For RQ2 (consistency between co-occurring components), we developed a consistency evaluation metric suitable for our case study. For RQ3 (impact of query formulation on content quality), we designed metrics to evaluate relevance, source quality, and safety measures across different query variations. Finally, for RQ4 (relationship between query sentiment and response sentiment), we created a set of queries with diverse sentiment valence and conducted manual annotation of the sentiment valence of AIO/FS to evaluate whether the sentiment of the outputs aligns with the query sentiment. Below, we detail our approach for each component of this research framework.

\subsection{Search Query Selection and Reformulation}

\paragraph{Initial Query selection.}
We curated a dataset of well-formed search queries by extracting queries relevant to our case study from a large public search query dataset in three iterative steps:
(1) keyword filtering using baby care- and pregnancy-related terms; (2) length-based filtering by retaining queries with at least three words and removing near-duplicates through pre-processing; (3) automated annotation with gpt-4o-mini to exclude queries unrelated to human baby care or pregnancy (e.g., about animals, celebrities, or non-informational content).

For the first step, we selected the queries related to baby care and pregnancy from the 10.4 million distinct Bing search queries in the Open Resource for Click Analysis in Search (ORCAS), which is a public click-based dataset associated with information retrieval tasks\footnote{This search query dataset contains no personally identifiable information {and has been filtered using a k-anonymity threshold to exclude infrequent queries. Offensive and pornographic content has been removed, meaning some extreme queries may be absent.}}~\cite{craswell2020orcas}.
Queries on baby care and pregnancy represent persistent everyday information needs rather than viral spikes, appearing in only 0.05\% of trend episodes in the GoogleTrendArchive dataset~\cite{urman2026trends}. In contrast, ORCAS captures the long tail of real user queries, making it better suited for studying realistic health information-seeking behavior.

For the first filtering step, we used the keywords ``baby" and ``babies" for the topic baby care and ``pregnant" and ``pregnancy" for the topic pregnancy. 
For the second step, we filtered for queries with a minimum length of 3 words, as 2-word queries (e.g., ``baby teeth," ``baby eyes") tend to be too vague. Additionally, we removed almost identical queries. To do this, we applied tokenization, stemming, and stopwords-removal to the queries, and then removed the queries that were duplicates after this pre-processing step. As a result, we obtained a total of 15,500 queries.

In the third step, we conducted further filtering of the queries to ensure we include only those related to human baby care and/or pregnancy. This was necessary to filter out queries related to (1) baby care/pregnancy of an animal or plant\footnote{{
We filtered out any plant-related queries that might use pregnancy-related terminology.}}; (2) information about pregnancy of a celebrity, fictional character, or a specific person; (3) a name, song, or item that simply contains the words ``baby" or ``pregnancy" but is not related to seeking advice or information on pregnancy. For this step, we relied on annotation using gpt-4o-mini. Following best practices for LLM-based annotation~\cite{baumann2025llmhacking}, we validated the model's performance on a subset of 50 manually annotated queries before applying it to our full dataset. The gpt-4o-mini classifications overlapped with our manual annotations 100\% of the time. The exact prompt we used can be found in the Appendix. In total, after this step, we retained 9516 queries relevant to questions about human baby care and/or pregnancy.

\paragraph{Question type classification (RQ1, 2, 3).}
To investigate the question type's impact on the appearance and content of AIO/FS, we obtained a balanced query dataset
over different question types and topics. Specifically, {from the 9,516 queries obtained in the previous filtering step}, we selected queries that fit the definition of well-formed questions from prior work \cite{chu2020ask}, i.e., queries that start with explicit question words \cite{wiki_defintion_question_words}. 
These queries were categorized into the following six groups of well-formed question types:
\begin{itemize}[leftmargin=3mm]
\item  \textbf{Binary Questions}: Questions that typically aim for a clear binary yes/no answer and start with one of the following words: ``do", ``does", ``did", ``is", ``are", ``was", ``were", ``can", ``could", ``will", ``would", ``should", ``may", ``might", ``shall", ``must", ``have", ``has", ``had", ``need". (note: question words in this category are further expanded compared with question words listed in  prior work \cite{chu2020ask}).
\item \textbf{Wh* Questions}: Questions asking about definitions, identifications, or directions, starting with words including like ``who", ``what", ``where", ``which", ``whose", and ``whom". 
\item  \textbf{When Questions}: Questions related to time, starting with the word ``when''.
\item  \textbf{How-to Questions}: Questions that start with ``how to" and typically seek steps or instructions for completing a process or task.
\item  \textbf{How + Adjective/Adverb Questions}: Questions that start with ``how" and are typically followed by an adjective or adverb, like ``how soon," ``how often,"  ``how many," ``how much," ``how far," etc., aiming to determine degree, quantity, or extent.
\item  \textbf{Why Questions}: Questions that start with ``why," aiming to find reasons, explanations, or causes behind something.
\end{itemize}
To ensure a balanced query dataset across different question types, we randomly sampled 100 queries per type per topic  (i.e., either baby care or pregnancy) when possible, except for the why question (as there are only 51 queries in total for baby care and pregnancy topic, as shown in Table~\ref{tab:well_formed_question_query_distribution}) and when question (as only 86 queries in total for pregnancy topic).
This resulted in a total of 1,037 queries.  
Table~\ref{tab:well_formed_question_query_distribution} shows a detailed overview of the well-formed question distribution by different query topics and question types.

\paragraph{Query sentiment reformulation (RQ4).}
To investigate how question sentiment affects AIO and FS appearance and content, we created sentiment variations of binary queries. We manually selected 157 binary queries {(23 baby care, 134 pregnancy) that could be naturally reformulated into neutral, positive, and negative versions using sentiment-bearing keywords.
Selection required that all three versions represent realistic search queries.}
For instance, the  neutral query ``Can infants have juice?'' becomes ``Is it safe/unsafe for infants to have juice?'' This process yielded 471 binary 
queries categorized as neutral/positive/negative.
\\\\
In total, through the steps above, we obtained 1,508 well-formed questions, which we used as input to the auditing pipeline described below.

\begin{table}[t]
  \centering
  \scriptsize
  \begin{tabular}{lp{0.7cm}p{0.8cm}p{0.8cm}p{0.8cm}p{0.8cm}p{0.6cm}}
\toprule
Topic & Binary  & How +adj/adv  & How to  & When & Wh*  & Why\\
\midrule
Babycare  & 117  & 209 & 140 & 222  & 122  &  28 \\
Pregnancy & 409  & 253 & 125 & 86 & 116  &23  \\
\bottomrule
\end{tabular}
  \caption{Summary statistics about the well-formed question query type distribution of baby care (Total n=838) and pregnancy (Total n=1012).}
  \label{tab:well_formed_question_query_distribution}
\end{table}

\subsection{Auditing Pipeline}

\paragraph{Data crawling.} We designed an automated pipeline to audit algorithmic behavior in response to 1,508 queries, using agents (i.e., specifically Selenium here) that simulate human searches from a fixed US-based IP address (set to New York City, NY). {To check the robustness of our results to location changes, we crawled data from five different locations. We did not find any significant differences across geographic location in terms of the appearance and content of FS and AIO (see Appendix for more details).
} Google SERPs were crawled between April 8-9, 2025 over the Google Chrome browser without logging into a Google account, so as to control for personalization factors. The SERPs were saved in HTML format during the crawling process, and AIO and FS responses were parsed and extracted from the HTML afterwards. This setup allows us to trace content and sources of AIO and FS responses, for analysis of how question type and sentiment affect results.

\paragraph{AIO visible versus suppressed.}
We observed that Google's AI Overviews (AIO) are sometimes present in the HTML of a results page but not shown to users{, and the corresponding result pages either show a banner ``$\star$ An AI Overview is not available for this search" on the very top of the page or nothing}.
The fact that Google restricts triggering visible AIO for certain queries has previously been described by~\cite{Williams2024whyAIOwrong}.
To capture this, we design a parser to identify both the visible AIOs and those embedded but hidden from view{. Specifically, we first verify AIO presence by checking for both a primary identifier element and a container element. Once confirmed, we then classify AIOs as visible or suppressed based on styling attributes and the absence of certain elements. We mark AIO as ``suppressed" when it lacks elements that typically accompany visible overviews and also lacks certain styling attributes, indicating that the AIO answer exists in the underlying HTML DOM structure but is deliberately hidden from users}.\footnote{{Code and data are available at \url{https://github.com/whocheers/Google-AIO-FS-Audit}.}}

\subsection{Manual Evaluation Metrics (RQ2, 3, 4)}

To evaluate the quality of outputs, we relied on manual annotation of the results. Building on prior work on information quality and contradictory text identification~\cite{es2024ragas,saad2023ares,li2023contradoc,de2008finding,wu2022topological}, we defined five core metrics tailored to our setting, i.e., \textit{consistency}, \textit{relevance}, \textit{safeguard}, \textit{domain category}, and \textit{sentiment valence}.

\paragraph{Consistency (RQ2).} Instead of adopting a strict logical definition of consistency, where prior work defines properties such as transitivity, commutativity, and negation invariance \cite{liu2024aligning}, we propose a looser definition that more closely matches human intuitions and facilitates annotation. This is inspired by prior work on contradiction detection in text \cite{li2023contradoc,de2008finding}. Given a query, consistency is defined between a pair of answers when there is no clear countervailing and contradictory statement/evidence. Namely, consistent answers provide equivalent or similar information, where one answer might offer complementary details without creating contradictions or express the same core information in different words.

In contrast, contradiction occurs when there is clear countervailing evidence between pairwise statements given a question. This can manifest through relatively obvious features (antonymy, negation, or numeric mismatches) or through complex differences in assertion structure, world-knowledge discrepancies, and lexical contrasts, as indicated in prior work \cite{de2008finding}. Prior work summarizes different types of contradiction \cite{li2023contradoc,wu2022topological}. We provide  a taxonomy of contradictions that are tailored to our research context as follows:
\begin{itemize}[leftmargin=3mm]
\item \textbf{Binary Contradictions}: Direct opposites where answers are mutually exclusive (yes/no, safe/unsafe, do/don’t, etc.) and cannot both be true or co-existing simultaneously (e.g., not dangerous vs. concerning). These are clear-cut disagreements, including the negation and antonymy types defined in previous work \cite{li2023contradoc,wu2022topological}.

\item  \textbf{Numeric Mismatches}: Different, but specific ranges, times, measurements or likelihoods that don't overlap and can create confusion; any  numerical or temporal or likelihood differences or mismatches, following a definition in prior work \cite{li2023contradoc}; 
 
 \item \textbf{Other Problematic Mismatches}: Different levels of relevance in addressing the question (e.g., FS and AIO address completely different aspects of the same topic, where one answer does not address the core question), different severity levels when risk is discussed, different pre-conditions (e.g., age limit/requirement, allergy history, approval/consultation from medical professionals), different degrees of certainty about effects,  or conflicts in causal relationships that don’t fit other categories.
 
 \end{itemize}

In addition to comparing the complete AIO and FS answers, we separately evaluated the consistency between their highlighted portions. This dual assessment simulates two distinct user behaviors: the ``quick scanner" who only reads highlighted text before drawing conclusions, and the ``thorough reader" who examines the entire content of both components. This approach was inspired by \citet{fernandez2024search}, who similarly modeled different user behavior modes (lazy versus diligent users) during health information seeking.

\paragraph{Relevance (RQ3a).} With this metric, we assess both the whole answer and the corresponding highlighted part in a pair-wise way. We are interested in assessing the highlighted part as previous work shows that highlighted text may affect people's trust in generative AI \cite{li2025human} and the highlighted part of a featured snippet may be irrelevant to the question \cite{zhao2019evaluation}. Specifically, we first examine the whole answer and assess whether the answer addresses all aspects of the question and aligns in topic/scope. We assign a \textit{high}, \textit{medium}, or \textit{low} relevance label, defined as:~\looseness=-1
\begin{itemize}[leftmargin=3mm]
\item  \textbf{High relevance}:  all aspects of the question/query are addressed, and the AIO/FS matches the query topic;
 
\item  \textbf{Medium relevance}: some, but not all, aspects of the question/query are addressed, and the AIO/FS matches the query topic; 
 
 \item \textbf{Low relevance}: no aspect of the question/query is addressed, or the query's topic is not addressed in AIO/FS.
 
 \end{itemize}

Then, for the highlighted part within each whole answer, we compare it side-by-side with the whole answer, and assess: ``Is the most relevant part in that answer addressing the question highlighted?" We assign a binary label ``Yes, the highlighted part is the most relevant part for addressing the question" or ``No, the highlighted part is NOT the most relevant part for addressing the question".

\paragraph{Safeguards (RQ3b).} Prior work found that AI answer engines (e.g., You.com, Perplexity.ai, Bing Chat) often use overtly confident language compared to traditional search engines (e.g., Google) \cite{venkit2024search} and that warning notes like ``Consult a doctor for medical advice" are important in the context of medical information \cite{scull2020dr}. Therefore, we assess whether AI overviews/Featured snippets include cues that caution the users to consult with a health professional on a given question. In connection to this, we come up with 3 categories:
\begin{itemize}[leftmargin=3mm]
\item  \textbf{Explicit safeguard cue:} Applies when it is asserted that consulting a medical professional is necessary or important, with cues like ``it’s crucial to consult ...'', ``it’s important to discuss with ...'', ``consult doctor as soon as ...'', etc.
 
\item  \textbf{Implicit safeguard cue:} Applies when consulting a medical professional is suggested as an option, without asserting the necessity or importance (with cues like ``may'',  ``as an option'', ``potentially'',  ``if needed'',  etc.);
 
 \item \textbf{No safeguard cue}: Applies when no suggestions to consult a medical professional were present.
 \end{itemize}
While we acknowledge there is no established ground truth for which queries should feature safeguards, this analysis represents the first systematic investigation of safeguard presence in generative search engines. It raises safeguards as an open challenge for the field while it also provides an initial step that foregrounds the issue for future research.

\paragraph{Source Category {and Credibility} (RQ3c).} Prior work has raised concerns regarding the variable quality of sources that the responses of generative AI search engines are based on, with evidence of overreliance on news, media and business \cite{li2024generative} websites. On the other hand, for featured snippets, Wikipedia has been reported to be the most common source based on an exploratory study using a large set of keywords \cite{strzelecki2019featured,strzelecki2020direct}. It is unclear, however, whether these findings apply for the specific topic of baby care and pregnancy, which serves as motivation for us to examine the top sources that are linked to in AIO and FS. Meanwhile, we also examine the top sources from the other ten blue links as a benchmark for comparison. For that, we first extract the corresponding websites (namely, domains) supporting AIO/FS and the ones from the other ten blue links of each SERP throughout the dataset. Then, we categorize each domain based on the FortiGuard Website categories\footnote{https://www.fortiguard.com/webfilter}, and further aggregate each domain's category and examine the category distribution overall. We opt for relying on FortiGuard as prior work \cite{vallina2020mis} found that FortiGuard’s categorization had the highest coverage of websites and the most accurate category labels among the providers of such categorization services. 

{Fortiguard categorizes websites but does not reliably indicate source quality: for instance, websites classified as Health can include both highly reliable scientific medical websites and sources such as Healthline that are effectively commercial media and contain unreliable, AI-generated content \cite{dupre2024wikipedia}. Thus, in a second step, we manually classified 144 unique domains (i.e., the top 10\% websites
most frequently mentioned in AIO, FS, and ten blue links
as: \textbf{High credibility} sources, including government and international health institutions (e.g., CDC, NHS, WHO), hospitals/clinics, medical associations (e.g., ACOG), and peer-reviewed scientific publications. \textbf{Medium/contextually-dependent credibility} sources, comprised of health-focused commercial media, healthcare service providers, general commercial media, non-profits, and reference websites like Wikipedia. \textbf{Low credibility} sources, including general and health-related e-commerce sites and social media platforms. The categorization was conducted by two of the authors who worked collaboratively to resolve all disagreements through discussion until consensus was reached.}

\paragraph{Sentiment Valence (RQ4).} 
This category was annotated only for the 471 well-formulated questions with different valence (see query reformulations subsection above). First, we extracted the subset of queries and corresponding SERPs where \textit{all} formulations of the same query (positive/negative/neutral) yielded AIOs and/or FS. There were 118 and 8 such query combinations for AIO and FS, respectively (i.e., 354 AIO queries and 24 FS queries in total). Disregarding the few FS queries, we classified AIO sentiment valence into one of the following categories:
\begin{itemize}[leftmargin=3mm]
\item  \textbf{Positive}: if the answer primarily discusses  benefits/positive aspects of the specific topic in the binary question.
\item  \textbf{Neutral}: if the answer equally addresses both benefits/positive aspects and harms/negative effects/cautions of the specific topic, or maintains a very neutral tone.
\item  \textbf{Negative}: if the answer primarily focuses on  harms/negative aspects of the specific topic in the binary question.
 \end{itemize}
 
We opted for manual assessment instead of using machine learning-based sentiment classification models as in \cite{li2024generative}, as this approach enabled us to capture the nuanced and complex differences in assertions, lexical contrasts, and subtle negation within statements. These linguistic features may alter the overall sentiment valence in ways that automated systems might overlook.

 \paragraph{Intercoder Agreement.} The consistency category was manually annotated by 3 coders, the corresponding Krippendorff's Alpha was 0.74 on 100 samples.
 All other categories were manually annotated by 2 coders, with the Krippendorff's Alphas as follows: 0.98 for relevance on 100 samples, 0.81 for sentiment annotation on 33 sample and 0.84 for safeguard annotation on 30 sample samples. Thus, across all categories we reached moderate (inconsistency) to high (all other categories) levels of agreement. Given that inconsistency involved 3 annotators and is, in general, a more nuanced metric, our level of agreement is acceptable for drawing further conclusions.
 All annotators discussed disagreements before annotating the remaining data.

 \section{Results}

\subsection{RQ1: Prevalence of AIO and FS}

\subsubsection{AIO/FS Appearance Overall.}

Overall, AIOs appear visibly\footnote{In addition to AIOs that are visible to the users, we find 157 AIO that, while present in the scraped HTML, are hidden from the users in the web interface. Hereafter, with the exception of ``Suppressed AIO" subsection, we discuss only the AIOs that were visible to the users.} in 84\% (1,272) of queries, while FSs appear in 32.5\% (490), with 22\% (322) of queries showing both elements co-occurring. Meanwhile, 88\% (429) of FS responses -- which we refer to as ``whole answers," i.e., the full response -- contain a highlighted section addressing the query, called the ``highlighted answer." Same terms are used for AIO responses, where 92.5\% (1,177) include a highlighted answer.

\subsubsection{AIO/FS Appearance across Question Type and Query Sentiment.} 
\begin{figure}[t!] 
\centering
  \includegraphics[width=\columnwidth]{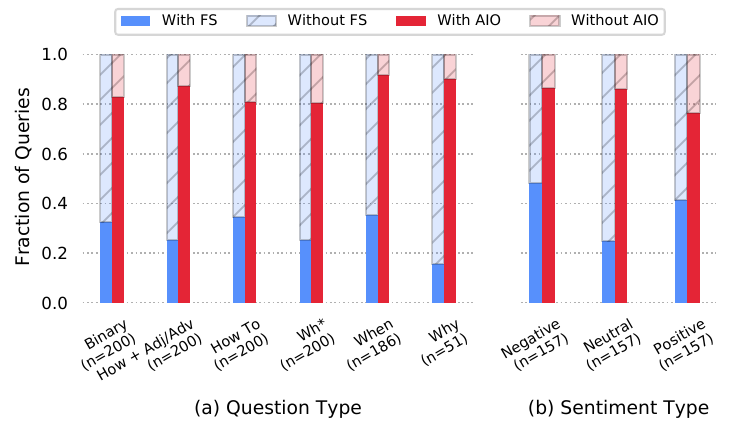}
  \caption{Fractional Appearance Distribution of AIO answer and FS answer by  question type and question sentiment. (Note: same $n$ for both bars in each pair here.)}
  \label{fig:AIO_FS_appearance_by_question_type_sentiment}
  \centering
\end{figure}

Chi-square tests show that question type is significantly associated with both FS answer ($p=0.015$) and AIO answers ($p<0.01$) frequencies.  FS answers are notably less common in ``Why" questions  (15.69\%) and more frequent in  ``When" questions (35.48\%), while AIOs maintain high appearance rates (80-92\%) across all question types, as shown in Figure~\ref{fig:AIO_FS_appearance_by_question_type_sentiment} (a).  

Query sentiment is also significantly (Chi-Square test $p<0.001$) associated with the frequency of appearance of FS. FS appears nearly twice as often for negative (48.41\%) compared to neutral (24.84\%) queries.  The difference is also statistically significant ($p=0.0265$) for AIO, with AIO appearing most often for negative (86.62\%) and neutral (85.99\%) queries, and slightly less often for positive ones (76.43\%) (as shown in Figure~\ref{fig:AIO_FS_appearance_by_question_type_sentiment}) (b).

\subsubsection{Suppressed AIO.}
\begin{figure}[t!] 
\centering
  \includegraphics[width=0.75\columnwidth]{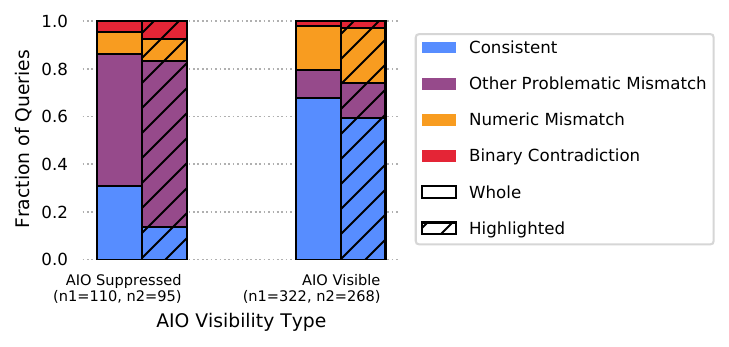}
  \caption{Fractional Consistency/Contradiction Distribution between AIO answer and FS answer by  AIO visibility.}
  \label{fig:contradition_by_AIO_visibility}
  \centering

\end{figure}

In addition to the AIO answers discussed in other Result sections that are visible to the users, we  also find 157 AIOs that are generated but not shown to the users. Thus, these AIOs were suppressed by Google. We have excluded these AIOs from our main analyses since they were not visible. However, we have conducted a comparison between the suppressed and the visible AIOs, and established that the suppressed AIOs were characterized by lower relevance than the visible ones and by higher inconsistency. Specifically, while only about 1\% of visible AIOs were of low relevance to the query, this was the case for 16\% of suppressed ones. In addition, only 31\% of suppressed whole AIO answers (versus 67\% visible whole AIO answers) were consistent with the FS when both were included in the SERP (see Figure \ref{fig:contradition_by_AIO_visibility} for details). Qualitative analyses of the suppressed AIOs showed that these elements often included either pro-life/anti-abortion or anti-vaccination content.

\begin{figure}[t!] 
\centering
  \includegraphics[width=\columnwidth]{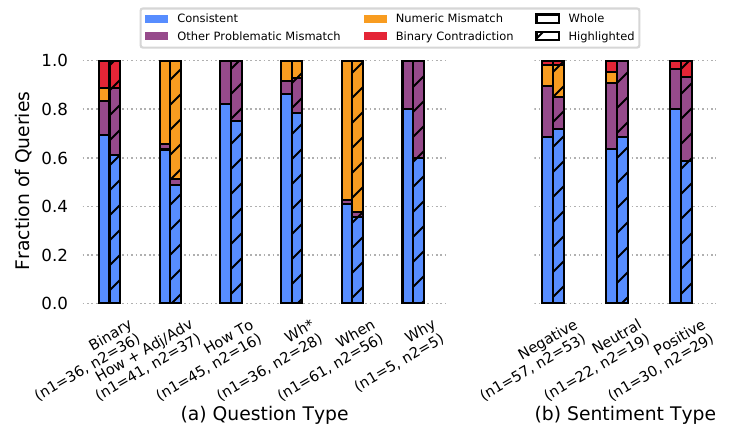}
  \caption{Fractional Consistency/Contradiction Distribution of AIO answer and FS answer pairs (including whole answer pairs and highlighted answer pairs) grouped by question type and sentiment.}
  \label{fig:contradition_by_question_type_and_sentiment}
  \centering
\end{figure}

\subsection{RQ2: Consistency of AIO and FS}
Consistency is evaluated only for 322 queries for which both AIO and FS are displayed on the SERPs simultaneously.

\subsubsection{Inconsistency of Whole/Highlighted Answers Overall.}
We conduct the analysis of inconsistency separately for the full AIO and FS answers, and the highlighted parts of the text within them for each of the 322 AIO/FS pairs. The shares of AIO/FS pairs with inconsistent responses are relatively high for both output types: 32.3\% for full answer pairs, and 40.7\% for highlighted text only pairs. Specifically, the fraction of the binary contradiction, numeric mismatch and other problematic mismatch is 1.8\%, 18.7\% and 11.8\% respectively for whole answers; and 2.6\%, 23.2\% and 14.9\% respectively for highlighted answers.

In Table~\ref{tab:contradictions} in the Appendix, we provide illustrative examples of each type of contradiction observed. {We also include a separate qualitative analysis of these examples in the Appendix. The examples collectively demonstrate that contradictory answers span from direct oppositions to subtle but important differences in conditions and risk assessments. What is more concerning is that seemingly low-stakes examples like feta cheese consumption, ice cream consumption, baby acne duration, and sleeping positions for pregnant women may have severe health consequences if readers miss critical risks or conditions.}

\subsubsection{Inconsistency Types vs. Question/Query Sentiment Types.}
We find significant differences in consistency labels across question types using Chi-square tests ($p<0.001$). Notably, only binary-type questions lead to binary contradictions between FS/AIO (11.1\% of such questions). ``How +adj/adv" questions lead to a high share of numeric mismatch contradictions: 34.2\% for full answer pairs, 48.7\% for highlighted text pairs. The shares of numeric mismatches are even higher for ``when" questions, 57.4\% and 62.5\%, respectively. As shown in Figure~\ref{fig:contradition_by_question_type_and_sentiment} (a), there are also relatively high shares of other problematic mismatches in the case of binary (13.9\% full; 27.8\% highlighted), ``how to" (17.8\%; 25\%), and ``why" questions (20\%; 40\%). 

In the case of query sentiment, we do not find a statistically significant difference in the consistency labels  ($p=0.052$). Yet, as shown in Figure~\ref{fig:contradition_by_question_type_and_sentiment} (b), the share of numeric mismatches is relatively high  for negatively-formulated queries (8.8\% for whole answer pairs and 4.6\% for highlight answer pairs) and appears as 9.1\% for the whole answer pairs of neutral queries and do not appear in the highlighted answer pairs of such set of queries; while positive query formulations did not yield numeric mismatches but did result in binary contradictions that are more prevalent than negative queries.

\subsection{RQ3a: Output Relevance} 
\subsubsection{Overall Relevance of AIO/FS.}
96.6\% of AIOs and 88.7\% of FS are rated as highly relevant to the  queries. Only about 1\% of AIO/FS showed low relevance to the queries. The detailed percentages for high/medium/low relevance are reported in Table~\ref{tab:relevance_labels_distribution}. Similarly, in the absolute majority of cases (95.5\% for AIO, 89.5\% for FS) the highlighted part of text in the AIO/FS is the one most relevant to the query. In Table~\ref{tab:low_relevance_example} in the Appendix, we present examples of low relevance ratings for both the full answer and the highlighted text.

\begin{table}[t]
  \centering
  \scriptsize
  \begin{tabular}{lp{1cm}p{1cm}p{1cm}p{1cm}p{1cm}}
\toprule
& \multicolumn{3}{c}{Whole answer relevance} & \multicolumn{2}{c}{  Highlighted part most relevant?} \\
\cmidrule(lr){2-4} \cmidrule(lr){5-6}
Type & High  & Medium  & Low  & Yes & No \\
\midrule
AIO & 96.6\%  & 2.6\% & 0.8\% & 95.5\%  & 4.5\% \\
FS & 88.7\% & 10.2\% & 1.1\% & 89.5\% & 10.5\% \\
\bottomrule
\end{tabular}
  \caption{Summary statistics about the fractions of  different  relevance level of whole answer from AIO and FS and the fraction of whether the paired highlighted part of the whole answer is the most relevant.}
  \label{tab:relevance_labels_distribution}
\end{table}

\subsubsection{AIO/FS Relevance by Question/Sentiment type.}
For AIO answers, Chi-square tests shows that there is no statistically significant difference in the relevance ratings across different question types and query sentiments ($p=0.66$ and $p=0.14$, respectively). For FS answers, there is a statistically significant ($p<0.01$) difference in the relevance rating for question types but not for the query sentiment ($p=0.31$). The detailed fractional distribution of AIO and FS across different question type and different sentiment can be seen from Figure \ref{fig:AIO_relevance_by_question_type_and_sentiment} and Figure\ref{fig:FS_relevance_by_question_type_and_sentiment} in the Appendix.

\subsection{RQ3b: Presence of Safeguard{s}}

\paragraph{Overall safeguard cue prevalence.} Overall, 89.4\% of  AIO and 92.8\% of FS answers do not contain any safeguard cues, such as the advice to ``consult a medical professional."
This is concerning, as prior work has emphasized the importance of including such disclaimers in health-related information\cite{scull2020dr}.

Among the answers that do include safeguard cues, AIO responses show a slightly higher prevalence than FS, which contrasts with findings from prior work~\cite{venkit2024search} suggesting that ``AI answer engines like You.com, Perplexity.ai, and Bing Chat tend to use overtly confident language compared to traditional search engines (i.e., Google Search)."  Specifically, 9\% of AIO answers contain an explicit safeguard cue and 2\% contain an implicit one. In contrast, 4\% of FS answers include an explicit safeguard cue, and 2\% include an implicit one.

\paragraph{Safeguard cues by question type and sentiment.} For AIO, Chi-square tests indicate a statistically significant difference in the presence of safeguard cues across different question types ($p < 0.001$), but not across query sentiments ($p = 0.095$).
For FS, no statistically significant differences is found in the presence of safeguard cues across either question type or sentiment ($p = 0.667$ and $p = 0.393$, respectively). While there's no significant difference among different question sentiments, the  fraction of safeguard cue within the set of queries formulated with different sentiment valence is higher than the set of queries formulated as different question types overall , eg., among the set of queries reformulated with sentiment, the fraction range of explicit safeguard cues is 13\% to 21\%  for AIO and 4\% to 10\% for FS, as shown in Figure \ref{fig:combined_AIO_FS_safeguard_by_question_type_and_sentiment} (b) in the Appendix; conversely, the fraction range of  explicit safeguard cue is 0 to 9\%  for AIO and 0 to 2\% for FS among the set of queries with different question types as shown in Figure~\ref{fig:combined_AIO_FS_safeguard_by_question_type_and_sentiment} (a).

\begin{figure}[t!] 
\centering
  \includegraphics[width=0.65\columnwidth]{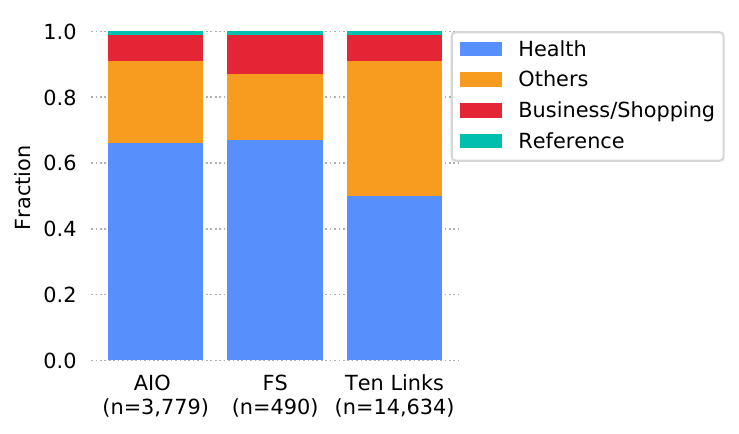}
  \caption{Fractional distribution of major   categories of domains sourcing AIO/FS answers and the ten blue links.}
  \label{fig:domain_categories_distribution_broad}
  \centering

\end{figure}

\subsection{RQ3c: Source Categories} 

\paragraph{Overall Categories of AIO/FS and ten other links.} In total, there were 3,779 domains linked to in AIOs, 490 in FS, and 14,634 appearing among regular search results. Out of those, there were 498, 121, and 1,415 unique domains respectively. These corresponded to 30, 15, and 41 unique categories as classified using Fortiguard.

Out of these unique categories, in our analyses we focus on four: ``Health and Wellness", ``Reference'', ``Business" and ``Shopping" category. Our reasoning for selecting these categories is that ``Health and Wellness'' category directly relates to the query topics of baby care and pregnancy; the ``Reference" category contains links to Wikipedia- the website, which prior work identified as the most common source domain for FS \cite{strzelecki2019featured,strzelecki2020direct}. We also include ``Business" and ``Shopping" categories as they are potentially problematic sources in the context of information on health topics like baby care and pregnancy. Information from categories beyond these four can be either reliable or problematic in relation to pregnancy and baby care, depending on the context. As shown in Figure \ref{fig:domain_categories_distribution_broad}, domains in the ``Health and Wellness" category constitute the highest fraction of sources linked to across AIO, FS, and regular search results. Using the Two Proportion Z Test, we find that its proportion among AIO and FS is significantly higher than among the ten blue links (both $p < 0.001$), suggesting that AIO/FS  prioritizes health-related websites for health queries like baby care and pregnancy. However, contrary to prior findings, the fraction of Wikipedia sources (i.e., ``Reference" category) is consistently low (approximately 1\%) across AIO, FS, and ten blue links. Most concerning, we found that FS answers source a significantly higher fraction from ``Shopping" and ``Business" categories ($p < 0.001$) than the ten blue links, potentially compromising the objectivity and quality of health information presented to users.

\begin{figure}[t!] 
\centering
  \includegraphics[width=\columnwidth]{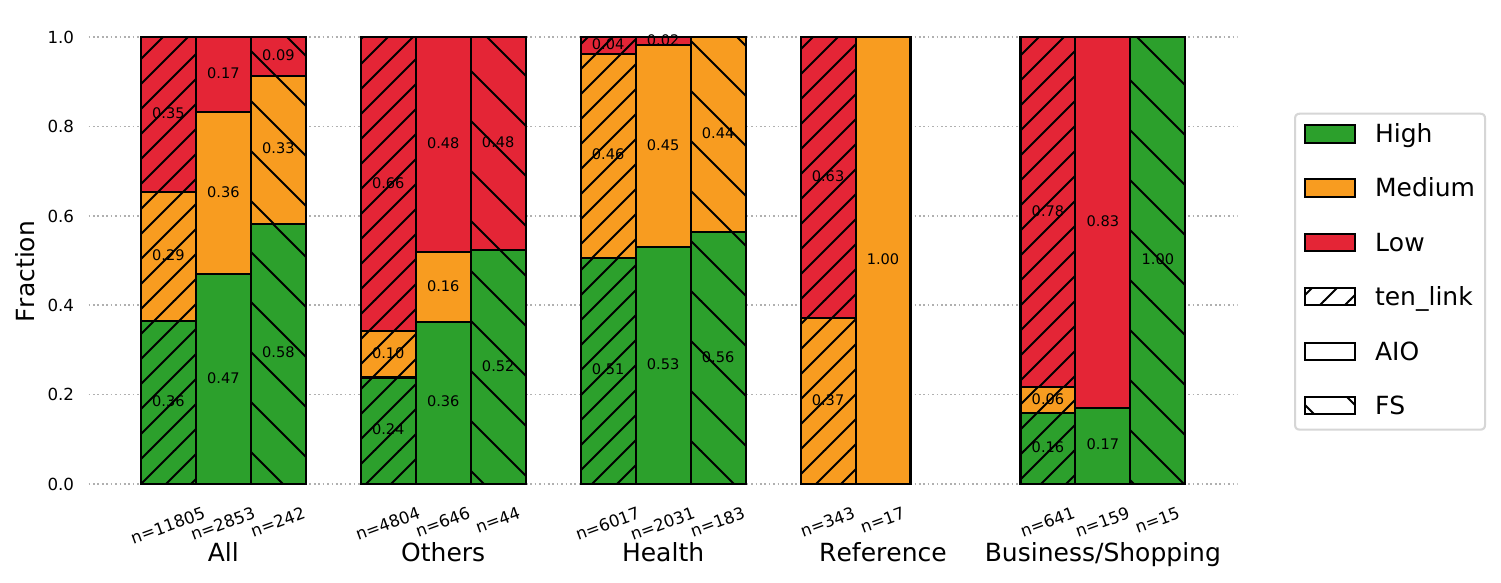}
  \caption{{Fractional distribution of source credibility for top 10\% domains in AIO/FS answers and the ten blue links.}}
  \label{fig:domain_credibility_distribution_broad}
  \centering
\end{figure}

\paragraph{{Source credibility.}}  {In Figure \ref{fig:domain_credibility_distribution_broad} we show the credibility distribution of sources among the top 10\% most frequently sourced domains (N=144) in either AIO/FS/ten blue links. Overall, sources with low credibility in the context of health and pregnancy, such as social media and e-commerce websites, are featured comparatively often, ranging from 9\% for FS to 17\% in AIO and 35\% for ten blue links. Low- and medium-credibility sources combined account for nearly half of all citations in AIO and FS. This raises concerns about the reliability of health information presented to users.}

 \subsection{RQ4: Query and Response Sentiment Valence} 
As mentioned in the methodology section, we labeled 118 sets of questions where positive, neutral, and negative versions all triggered AIO responses, resulting in 354 pairs of questions and their corresponding AIO responses labeled as \textit{positive}, \textit{neutral}, or \textit{negative}.

Based on these 354 pairs of search queries and their corresponding answers, a chi-squared test indicates no statistically significant association between question and answer sentiment ($\chi^2(4) = 8.24$, $p = 0.08$). Namely, in our research setting. when controlling for both the query topic and query type, we do not observe the ``confirmation bias" claimed in prior work where these factors were not controlled \cite{li2024generative,venkit2024search}.

\section{Discussion}

{Over the past decade, Google's SERP has evolved from  ten blue links to sophisticated search features including top stories, knowledge panels, and, more recently, FS, and AIO. Correspondingly, user behavior has shifted from traditional search patterns (i.e., clicking through multiple vertical links) toward engaging with direct answers that address search intentions immediately, reducing search effort and improving user satisfaction \cite{wu2020providing}. This evolution reflects a consistent trend: aggregating results from heterogeneous sources within a single interface to directly address searcher intent \cite{oliveira2023evolution}. While specific SERP features may change, future developments will likely include such as conversational functions (already adopted by other search engines~\cite{capra2023does}), and all these will most likely follow this same trajectory of providing immediate answers. As new search features emerge, our research methodology for examining how query characteristics affect SERP results through six quality evaluation metrics will therefore remain relevant and transferable to other high-stakes domains.}

{Within this evolving landscape, our analysis of baby care and pregnancy searches reveals several concerning implications.} The high prevalence of AIO at the top of the SERP highlights their growing role in shaping how people access and consume information. As AIOs increasingly mediate knowledge access, their quality becomes crucial across other domains like legal advice and political information. Importantly, both query sentiment and question type are significantly associated with the presence of AIOs and FSs, suggesting that seemingly subtle variations in how users formulate their questions may shape which information components they see, especially for featured snippets. This has direct implications for user equity and information exposure across varied demographic or informational contexts, where diverse groups may formulate the same questions differently regarding sentiment and question types. 

One of the most concerning findings is the inconsistency between AIO and FS answers when they co-occur. \textit{Binary contradictions}, such as the conflicting guidance on feta cheese consumption during pregnancy  (Figure~\ref{fig:figure1}), are particularly alarming and highlight the potential for user confusion and harmful decision-making in high-stakes contexts like health, which require clear and consistent guidance. {This feta cheese example illustrates how seemingly low-stakes queries can mask serious health risks: feta carries listeria infection risks that can trigger miscarriage, demonstrating the difficulty of clearly delineating high-stakes contexts in health information. We conceptualize harm along multiple risk dimensions: missing critical information, exposure to false guidance, and overconfident or conflicting responses. Our observed contradiction rate of 32.3\% - 40.7\% (depending on whole  or highlighted answer pairs) when AIO and FS co-occur represents systematic exposure to conflicting authoritative-seeming information that compounds risk, particularly for vulnerable populations like pregnant users who rely heavily on search  for health guidance.}

{From a user experience perspective, encountering contradictory information between two prominently displayed, algorithmically-endorsed sources disrupts interaction patterns by prompting users to conduct additional searches to resolve conflicting guidance, creating friction in information-seeking workflows and potentially delaying critical decisions. These inconsistencies may also erode user trust of GenAI search system, as well as may directly affect users' knowledge gathering and decision-making processes and create uneven experiences that disadvantage users lacking  digital literacy to recognize when additional verification is needed. User studies examining how conflicting AI-generated information affects user search behaviors, perception of trust, knowledge acquisition and decision-making represent important future research directions.}

Other contradiction types like \textit{numerical mismatches} regarding recommended timeframes or dosage amounts also pose potential risks. These issues are further complicated by question type, which is significantly related to consistency. This suggests that how a query is framed (e.g., asking for instructions via ``how to" vs. explanations via ``why"/``what") may directly impact the coherence of the information presented, raising broader concerns about information reliability in fields such as law or finance.

Source category analysis shows both encouraging and troubling trends. AIO/FS answers are more likely to source from health-related websites than traditional links, which is reassuring. 
However, FSs disproportionately rely on business and shopping websites.  This raises concerns, especially since FSs typically derive their answers from a single source. {Meanwhile, manual review of top 10\% domains further reveals that low- and medium-credibility sources comprise nearly half of all AIO/FS citations, raising additional concerns given the considerable fraction of inconsistent answers between AIO and FS, as this combination may amplify potential health risks for users.}  As question type also significantly influences the source category distribution for both AIOs and FSs, different query formulation styles may expose users to different categories of source material (which may come with varying quality levels). This may be particularly problematic in areas where commercial incentives could skew the credibility of information, such as political advertising or financial products.

The presence of safeguard cues, such as ``consult a doctor'', is limited in both AIO and FS answers. While AIOs include such warnings more often, contrasting with findings from earlier studies about AI search engines' linguistic overconfidence~\cite{venkit2024search}, the overall low frequency is concerning given the sensitive nature of health-related information. Moreover, safeguard inclusion is significantly associated with question type for AIOs, indicating that some query formulations are more likely to trigger cautionary responses than others. This observation underscores the importance of auditing how search engines calibrate their tone and risk communication across various domains, including those where disclaimers or warnings might be commonly necessary (e.g., medical or legal advice).

We found no evidence of confirmation bias in AIO responses -- operationalized as overlap between query and AIO sentiment -- when controlling for query topic and question type. Thus, at least within our case study of pregnancy and baby care, Google’s systems are either applying effective mitigation strategies for sentiment's impact or the topic space itself supports more consensus-based answers. Notably, while query sentiment is significantly related to which components (AIO/FS) appear, it is not significantly associated with the relevance, safeguard, source category, or sentiment of the responses themselves.

\paragraph{Limitations.} Our study has three main limitations. First, data collection is restricted to the US and English-language results. Second, our analysis represents a single point in time rather than tracking changes longitudinally. Third, we focus exclusively on Google -- a choice explained by its dominant market position both in the US and globally over the past decade~\cite{statista-search-usa,statcounter-search-global}.

\section*{Acknowledgments}
We thank Ronald E. Robertson, Kaicheng Yang, Piotr Sapieżyński, and Jeffrey Gleason for their valuable feedback on this work. We also thank everyone in the Social Computing Group at the University of Zurich and the anonymous reviewers for their helpful comments and suggestions.

The work of Aleksandra Urman, Elsa Lichtenegger, and Aniko Hannak was supported by Swiss National Science Foundation (SNSF) Grant 215354; Joachim Baumann is supported by SNSF grant P500-2\_235328; and the work of Robin Forsberg was supported by the Kone Foundation.

\bibliography{aaai2026}

\clearpage

\appendix

\subsection{Paper Checklist}

\begin{enumerate}

\item For most authors...
\begin{enumerate}
    \item  Would answering this research question advance science without violating social contracts, such as violating privacy norms, perpetuating unfair profiling, exacerbating the socio-economic divide, or implying disrespect to societies or cultures?
   \answerYes{Yes}
  \item Do your main claims in the abstract and introduction accurately reflect the paper's contributions and scope?
  \answerYes{Yes}
   \item Do you clarify how the proposed methodological approach is appropriate for the claims made? 
   \answerYes{Yes, see the methodology section}
   \item Do you clarify what are possible artifacts in the data used, given population-specific distributions?
   \answerNA{NA}
  \item Did you describe the limitations of your work?
   \answerYes{Yes, see the Limitations subsection in the Methodology section}
  \item Did you discuss any potential negative societal impacts of your work?
 \answerNA{NA}
      \item Did you discuss any potential misuse of your work?
\answerNA{NA}
    \item Did you describe steps taken to prevent or mitigate potential negative outcomes of the research, such as data and model documentation, data anonymization, responsible release, access control, and the reproducibility of findings?
 \answerNA{NA}
  \item Have you read the ethics review guidelines and ensured that your paper conforms to them?
    \answerYes{Yes}
\end{enumerate}

\item Additionally, if your study involves hypotheses testing...
\begin{enumerate}
  \item Did you clearly state the assumptions underlying all theoretical results?
    \answerNA{NA}
  \item Have you provided justifications for all theoretical results?
     \answerNA{NA}
  \item Did you discuss competing hypotheses or theories that might challenge or complement your theoretical results?
    \answerNA{NA}
  \item Have you considered alternative mechanisms or explanations that might account for the same outcomes observed in your study?
 \answerNA{NA}
  \item Did you address potential biases or limitations in your theoretical framework?
   \answerNA{NA}
  \item Have you related your theoretical results to the existing literature in social science?
 \answerNA{NA}
  \item Did you discuss the implications of your theoretical results for policy, practice, or further research in the social science domain?
  \answerNA{NA}
\end{enumerate}

\item Additionally, if you are including theoretical proofs...
\begin{enumerate}
  \item Did you state the full set of assumptions of all theoretical results?
  \answerNA{NA}
	\item Did you include complete proofs of all theoretical results?
    \answerNA{NA}
\end{enumerate}

\item Additionally, if you ran machine learning experiments...
\begin{enumerate}
  \item Did you include the code, data, and instructions needed to reproduce the main experimental results (either in the supplemental material or as a URL)?
     \answerNA{NA}
  \item Did you specify all the training details (e.g., data splits, hyperparameters, how they were chosen)?
     \answerNA{NA}
     \item Did you report error bars (e.g., with respect to the random seed after running experiments multiple times)?
  \answerNA{NA}
	\item Did you include the total amount of compute and the type of resources used (e.g., type of GPUs, internal cluster, or cloud provider)?
 \answerNA{NA}
     \item Do you justify how the proposed evaluation is sufficient and appropriate to the claims made? 
 \answerNA{NA}
     \item Do you discuss what is ``the cost`` of misclassification and fault (in)tolerance?
   \answerNA{NA}
  
\end{enumerate}

\item Additionally, if you are using existing assets (e.g., code, data, models) or curating/releasing new assets, \textbf{without compromising anonymity}...
\begin{enumerate}
  \item If your work uses existing assets, did you cite the creators?
     \answerYes{Yes}
  \item Did you mention the license of the assets?
    \answerNA{NA}
  \item Did you include any new assets in the supplemental material or as a URL?
    \answerNo{No, because we plan to publish our code and our crawled data for further research in our community upon publication of our work.}
  \item Did you discuss whether and how consent was obtained from people whose data you're using/curating?
  \answerYes{Yes, as mentioned in methodology, our search queries dataset are sourced from a public dataset, which is available publicly and free of charge to facilitate the research in this field.} 
  \item Did you discuss whether the data you are using/curating contains personally identifiable information or offensive content?
    \answerYes{Yes}
\item If you are curating or releasing new datasets, did you discuss how you intend to make your datasets FAIR?
   \answerNA{NA}
\item If you are curating or releasing new datasets, did you create a Datasheet for the Dataset? 
   \answerNA{NA}
\end{enumerate}
\item Additionally, if you used crowdsourcing or conducted research with human subjects, \textbf{without compromising anonymity}...
\begin{enumerate}
  \item Did you include the full text of instructions given to participants and screenshots?
   \answerNA{NA}
  \item Did you describe any potential participant risks, with mentions of Institutional Review Board (IRB) approvals?
 \answerNA{NA}
  \item Did you include the estimated hourly wage paid to participants and the total amount spent on participant compensation?
   \answerNA{NA}
   \item Did you discuss how data is stored, shared, and deidentified?
  \answerNA{NA}
\end{enumerate}

\end{enumerate}

\clearpage

\section*{Appendix}
\addcontentsline{toc}{section}{Appendix}

\section{{Geographical location robustness}}
{To investigate whether the geographic location of the user might affect the appearance and content of Featured Snippets (FS) and AI Overviews (AIO), we collected SERPs from five different US cities (New York, NY; Kansas City, MO; San Francisco, CA; Houston, TX; Madison, WI).
This additional data collection was conducted during September 2-9, 2024, using a dataset of 528 baby care and pregnancy queries and their variations. The queries were obtained from (1) a researcher who was pregnant and became a new mother, and (2) observational trace data collected from a panel of US residents that includes their web browsing into Google Search.\footnote{{While this data collection was approved by the institutional IRB, we cannot make the search query dataset from this robustness experiment publicly available because the nature of the data precludes deidentification; therefore, only aggregated results are presented in this manuscript.}}
Since proxy servers and VPNs may not accurately reflect location in Google's search engine result pages (SERPs) during our manual examination, we used a geospoofing method by injecting precise latitude and longitude coordinates into automated scripts, similar to methods used in previous work~\cite{jung2025algorithmic}.
}

{
\textbf{We found no significant differences in either appearance or content across different geo-locations tested.}
We found nearly identical fractions of FS and AIO across locations, with FS appearing in 71.4\%-73.9\% of results and AIO in 90.7\%-92.6\% of results. Chi-square tests revealed no statistically significant differences between location and FS/AIO appearance ($p = 0.50$ and $0.84$ respectively).}

{To investigate content differences, we randomly sampled 100 queries from the 528. For FS content comparison, we used exact text matching since feature snippets are typically extracted verbatim from webpages and remain stable. For AIO content, which is more volatile and commonly rephrased even for identical searches, we measured consistency across locations using two annotators who compared each location's AIO against a reference location and assigned consistency labels (Krippendorff's Alpha = 1.0 on 10 samples). Chi-square tests showed no statistically significant differences in FS or AIO content across locations ($p = 0.37$ and $0.33$ respectively).}

\section{{Qualitative analysis of inconsistent answers}}

{Among these three types of inconsistent answers, the first category, \textbf{binary contradiction}, is relatively rare but potentially the most problematic. These involve mutually exclusive statements that may mislead users when making decisions, as shown in Examples 1--2 in Table~\ref{tab:contradictions}. For example, Example 2 shows contradictory guidance about feta cheese consumption during pregnancy, with one response categorizing it as ``unsafe'' while the other includes it among ``safe'' dairy products. This contradiction is particularly concerning since feta cheese made from unpasteurized milk may contain Listeria bacteria, which can trigger severe complications including miscarriage.} 

{\textbf{Numerical mismatches} represent another category of contradictions, where paired answers provide different time frames or quantities (e.g., age, duration, or frequency) for the same health-related queries, as shown in Examples 3--4. For instance, Example 4 shows a temporal mismatch about baby acne duration.}

{The third category, broadly termed \textbf{``other problematic mismatches,''} encompasses various discrepancies in specific conditions (e.g., age limits, health conditions, professional approval), risk levels, or even interpretation of the question itself, as shown in Examples 5--9. For example, Example 6 reveals conditional differences regarding ice cream consumption during pregnancy, with one answer detailing specific safety criteria (i.e., made from pasteurized milk with no raw egg) while the other offers general dietary guidance. Similarly, Example 7 shows how both answers recommend left-side sleeping for pregnant women, but one emphasizes significantly higher risks associated with alternative positions, including stillbirth and preeclampsia.}

 \clearpage
\section{Implementation details}
\label{app:implementation}

We used the following prompt for the last step of the query filtering with the model \texttt{gpt-4o-mini-2024-07-18}:

\begin{tcolorbox}[colframe=gray!90, colback=gray!20, 
title=System prompt, fonttitle=\bfseries, 
boxrule=0.5pt, left=5pt, right=5pt, top=5pt, bottom=5pt, boxsep=5pt]

You are an assistant to determine whether a query is related to daily life baby care or pregnancy. Exclude queries that refer to:

1. Baby/pregnancy of an animal or plant

2. Baby/pregnancy of a celebrity or fictional character or a specific person

3. A name, song, or item that simply contains the words ``baby" or ``pregnancy"

For such queries, return Label: No. Otherwise, return Label: Yes if the query is genuinely related to human baby care or pregnancy.
Examples:
Query: ``did Baby Yoda die"
Label: No
Explanation: ``Baby Yoda" is a fictional character.

Query: ``are baby carrots healthy"
Label: No
Explanation: ``Baby carrots" refer to a vegetable, not human babies.

Query: ``what is the word for baby in Irish"
Label: No
Explanation: This is about translation, not baby care or pregnancy.

Query: ``is Jack pregnant"
Label: No
Explanation: This refers to a specific individual's pregnancy, not a general baby care or pregnancy topic.

Query: ``babies and honey"
Label: Yes
Explanation: This relates to baby care and the safety of honey for babies.

Query: ``Zoloft while pregnant"
Label: Yes
Explanation: This concerns pregnancy and the medication Zoloft.

Query: ``is a fetus a baby"
Label: Yes
Explanation: This discusses the definition of a baby in relation to pregnancy.
\end{tcolorbox}

\begin{tcolorbox}[colframe=gray!90, colback=gray!20, 
title=User prompt, fonttitle=\bfseries, 
boxrule=0.5pt, left=5pt, right=5pt, top=5pt, bottom=5pt, boxsep=5pt]

    Determine if a query is related to daily life baby care or pregnancy.
    
    query: \{query\}

    Return `Yes' if the query is related to daily life baby care or pregnancy, or `No' if it is not.

\end{tcolorbox}

\section{Additional experimental results}
\label{app:additional_results}

\paragraph{AIO/FS Source Category by Question/Sentiment type.}

For AIO answers, Chi-square tests shows that there is a statistically significant difference in the prevalence of different source categories across different question types ($p<0.001$) but not query sentiments ($p=0.68$). Similarly, for FS answers, there is a statistically significant difference across different question types ($p=0.017$)  but not query sentiments ($p=0.52$). As seen in Figure~\ref{fig:domain_categories_distribution_by_question_type_sentiment} (a), we find ``When" question queries resulted in the highest fraction of websites from ``Business/Shopping" (10.8\% to 22.7\%) and the lowest fraction of websites from ``Health and Wellness" categories (45.5\% to 63.1\%), while ``Why" questions sourced the lowest fraction of websites from ``Business/Shopping" (0\% to 1.5\%). Conversely, the distributions of different domain categories among different query sentiments are very similar, as shown in Figure~\ref{fig:domain_categories_distribution_by_question_type_sentiment} (b).

Table~\ref{tab:top_domain_cateogry} provides a more detailed overview of identified domain types across SERP components and query types.

\begin{figure}[t!] 
\centering
  \includegraphics[width=\columnwidth]{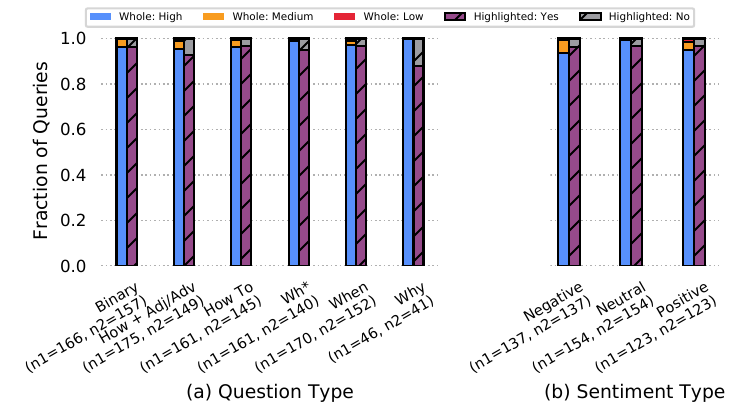}
  \caption{Fractional Relevance Label Distribution of AIO answer by different question type and different sentiment.}
  \label{fig:AIO_relevance_by_question_type_and_sentiment}
  \centering
\end{figure}

\begin{figure}[t!] 
\centering
  \includegraphics[width=\columnwidth]{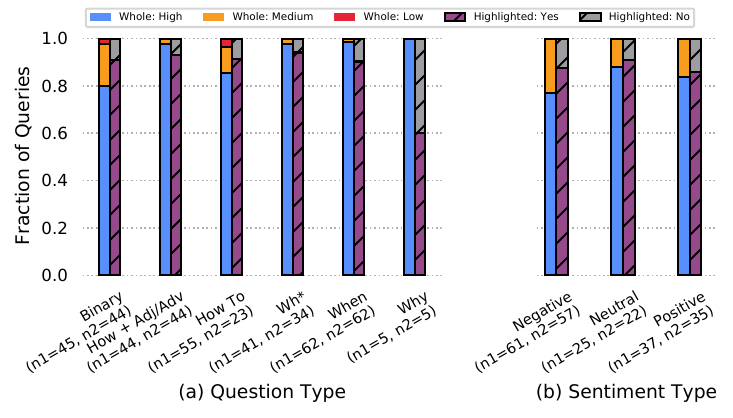}
  \caption{Fractional Relevance Label Distribution of FS answer by different question type and different sentiment.}
  \label{fig:FS_relevance_by_question_type_and_sentiment}
  \centering
\end{figure}

\begin{table*}[t]
\centering

\begin{tabular}{|p{0.7cm}|p{1.9cm}|p{3cm}|p{5.5cm}|p{5.5cm}|}
\hline
\textbf{Index} & \textbf{Contradiction Type} & \textbf{Question} & \textbf{AIO Answer} & \textbf{FS Answer} \\
\hline
1 & Binary contradiction & \textit{is it safe for you to have lunch meat while pregnant?} & \textit{lunch meat is generally considered \textbf{safe} to eat when pregnant, it's recommended to heat it [...]} & \textit{... Deli meat \textbf{isn't safe} for pregnant women because [...] can be much worse for pregnant women and their babies.} \\
\hline
2 & Binary contradiction & \textit{is it unsafe for me to eat feta when pregnant?} & \textit{Yes, feta cheese can be \textbf{unsafe} to eat during pregnancy ... can be dangerous for pregnant women and their fetuses.} & \textit{You're \textbf{safe} to eat some milk and dairy foods, including..., such as cottage cheese, mozzarella, feta,...} \\
\hline
3 & Numerical mismatch & \textit{when can babies start eating cereal?} & \textit{Most babies can start eating cereal around \textbf{6 months} of age ...} & \textit{... By the time they are \textbf{7 or 8 months}, your child can eat various foods ... include: Infant cereals} \\
\hline
4 & Numerical mismatch & \textit{how long does baby acne last?} & \textit{Baby acne, ... typically clears up on its own \textbf{within a few weeks or months} ...} & \textit{Baby acne could clear up on its own within \textbf{a few days to a couple of weeks}. The timeline \textbf{could be faster} ...} \\
\hline
5 & Other problematic mismatch & \textit{can baby sleep with pacifier?} & \textit{Yes, it is generally safe for babies to sleep with a pacifier. Benefits: Soothing: Sucking on a pacifier can soothe ... may reduce the risk of Sudden Infant Death Syndrome (SIDS)....} & \textit{... a newborn baby can start sleeping with a pacifier \textbf{in certain situations}. ... bottle-fed babies can sleep with a pacifier from birth, \textbf{breastfed infants should only} sleep with a pacifier \textbf{once they are at least 3 to 4 weeks old and have settled into a breastfeeding routine} with no latching issues.} \\
\hline
6 & Other problematic mismatch & \textit{can pregnant women eat ice cream?} & \textit{Yes, pregnant women can generally eat ice cream, \textbf{as long as it meets the following criteria: Pasteurized}: The ice cream should .... \textbf{No raw eggs}: Avoid ice creams that contain raw eggs, as they can carry the risk of salmonella poisoning...} & \textit{Enjoy ice cream as you would other sugary treats... come up with a healthy dietary pattern that allows room for your favorite foods to promote a healthy pregnancy.} \\
\hline
7 & Other problematic mismatch & \textit{what side should pregnant women sleep on?} & \textit{... recommend that pregnant women sleep on their left side. Reasons for Sleeping on the Left Side: Improved blood flow: .... allowing for optimal blood circulation.} & \textit{... advise pregnant women to sleep on their left side ... back and right-side sleeping with a \textbf{higher risk of stillbirth, reduced fetal growth, low birth weight, and preeclampsia, a life-threatening high blood pressure disorder} that affects the mother ...} \\
\hline
8 & Other problematic mismatch & \textit{can a pregnant woman take alka seltzer?} & \textit{No, pregnant women should generally avoid taking Alka-Seltzer. Alka-Seltzer contains aspirin, ... can \textbf{increase the risk of: Premature birth, Low amniotic fluid levels, Birth defects, and Kidney problems.}} & \textit{... Alka Seltzer is an over-the-counter medicine frequently used to treat heartburn. However, an active ingredient in many ... be avoided by pregnant patients unless prescribed by your provider to treat ...} \\
\hline
9 & Other problematic mismatch & \textit{how to bring up a newborn baby?} & \textit{Raising a newborn involves basic care like feeding ... providing a nurturing environment. It's crucial to prioritize sleep, seek support, and stock up on essentials ....} & \textit{To pick up your baby, slide one hand under their head and neck and the other hand under their bottom. Bend your knees to ... bring them close to your chest as you straighten your legs again.} \\
\hline
\end{tabular}

\caption{Examples of contradictory answers between AIO and FS.}
\label{tab:contradictions}
\end{table*}

\begin{table*}[t]
\tiny
\centering

\begin{tabular}{|p{0.7cm}|p{1.9cm}|p{3cm}|p{5.5cm}|p{5.5cm}|}
\hline
\textbf{Index} & \textbf{Label} & \textbf{Question} & \textbf{Whole Answer or Highlight Part} & \textbf{Explanation} \\
\hline
1 & low relevance & \textit{How to help baby acne?} & How to Help Baby Acne: Understanding
Baby Acne. Baby acne, also known as neonatal acne, ...typically
appears as small, red pimples on the face, neck, and
chest. & This whole answer is assigned a low relevance rating because
it fails to address the ”how to help” aspect of the question,
instead focusing on defining baby acne.

\\
\hline
2 & low relevance & Is it safe for a woman to
get pregnant while breastfeeding? & Yes,
it is possible to get pregnant while breastfeeding, even if
you haven’t had your period return after giving birth. While
breastfeeding can delay ovulation and reduce the chances
of pregnancy, it’s not a reliable form of birth control, especially
as the baby gets older.... not a foolproof method of
preventing pregnancy. & This whole answer receives a low relevance
rating because it primarily addresses the possibility of
getting pregnant while breastfeeding rather than the safety
concerns, which was the central focus of the question.\\
\hline
3 & No, highlighted part is not the most  relevant & Is it unsafe for you
to eat brie when pregnant? &...should avoid ... \textbf{all
mould-ripened soft cheeses with a white coating on the outside,
such as}......  & The highlighted part is: ”all
mould-ripened soft cheeses with a white coating on the outside,
such as...”. We mark this as not being the most relevant
highlight because it omits the crucial recommendation
that pregnant women ”should avoid” these cheeses, which
directly answers the safety question posed. \\

\hline
\end{tabular}

\caption{Illustrative Example of whole answers annotated as ``low relevance" or highlight part is NOT the most relevant.}
\label{tab:low_relevance_example}
\end{table*}

\begin{figure}[t!] 
\centering
  \includegraphics[width=\columnwidth]{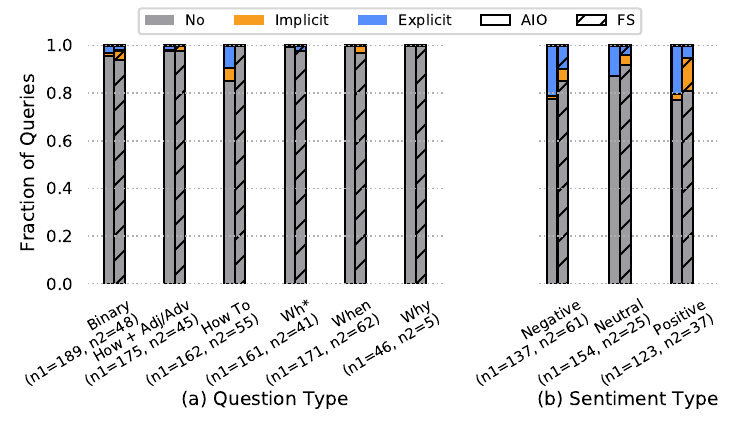}
  \caption{Fractional Safeguard Label Distribution of AIO and FS answer by  question type \& different sentiment.}
  \label{fig:combined_AIO_FS_safeguard_by_question_type_and_sentiment}
  \centering
\end{figure}

\begin{figure}[t!] 
\centering
  \includegraphics[width=\columnwidth]{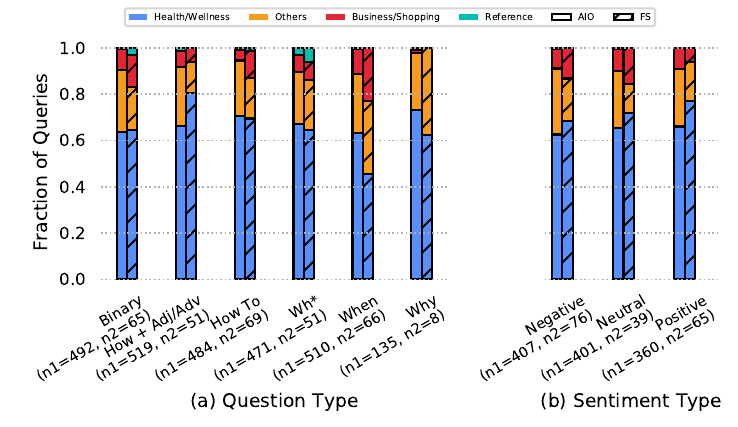}
  \caption{Fractional distribution of major categories of domains referenced in AIO/FS answers across different question types and query sentiments.}
  \label{fig:domain_categories_distribution_by_question_type_sentiment}
  \centering
\end{figure}


\begin{table*}[t]
\tiny
\centering
\begin{tabular}{p{2cm}|p{3.4cm}|p{2cm}|p{3.4cm}|p{2cm}|p{3.4cm}}


\toprule
                                      \multicolumn{1}{c}{AIO} && \multicolumn{1}{c}{FS} && \multicolumn{2}{c}{Ten Blue Links} \\
                                      \midrule
                     Category (Raw Count, Fraction) &                                                                                        Top Domains (Fraction of Each Top Domain) &                     Category (Raw Count, Fraction) &                                                                                                     Top Domains (Fraction of Each Top Domain) &                      Category (Raw Count, Fraction) &                                                                                           Top Domains (Fraction of Each Top Domain) \\
\midrule
               Health and Wellness (n = 2493, 0.66) &           healthline.com (0.138); nhs.uk (0.105); mayoclinic.org (0.089)  &                Health and Wellness (n = 329, 0.67) &                        nhs.uk (0.149); healthline.com (0.143); mayoclinic.org (0.082) &                Health and Wellness (n = 7342, 0.50) &              nhs.uk (0.105); healthline.com (0.083); mayoclinic.org (0.063) \\
             Society and Lifestyles (n = 293, 0.08) &         whattoexpect.com (0.420); parents.com (0.259); thebump.com (0.259) &                            \textbf{Business (n = 47, 0.10)} &                           nhsinform.scot (0.319); solidstarts.com (0.149); pampers.com (0.128) &      Newsgroups and Message Boards (n = 1808, 0.12) &                                reddit.com (0.998); boards.weddingbee.com (0.001); knittingparadise.com (0.001) \\
                           \textbf{Business (n = 263, 0.07)} &                 pampers.com (0.205); solidstarts.com (0.129); nhsinform.scot (0.103) &  Government and Legal Organizations (n = 28, 0.06) &                                        cdc.gov (0.464); fda.gov (0.250); ncbi.nlm.nih.gov (0.071) &             Society and Lifestyles (n = 1074, 0.07) &      thebump.com (0.304); whattoexpect.com (0.295); parents.com (0.250)\\
 Government and Legal Organizations (n = 207, 0.05) &                  ncbi.nlm.nih.gov (0.242); cdc.gov (0.150); pmc.ncbi.nlm.nih.gov (0.145) &              Society and Lifestyles (n = 19, 0.04) &                                                                            thebump.com (0.526); parents.com (0.316); whattoexpect.com (0.158) &                           \textbf{ Business (n = 968, 0.07)} &                    pampers.com (0.154); flo.health (0.075); nhsinform.scot (0.075) \\
                          Education (n = 116, 0.03) &  acog.org (0.328); health.harvard.edu (0.069); takingcarababies.com (0.043) &                           Education (n = 16, 0.03) &  acog.org (0.625); healthcare.utah.edu (0.062); healthyliving.extension.wisc.edu (0.062) &  Government and Legal Organizations (n = 874, 0.06) &           pmc.ncbi.nlm.nih.gov (0.316); ncbi.nlm.nih.gov (0.160); cdc.gov (0.156) \\
              Information Technology (n = 68, 0.02) &            utswmed.org (0.632); drsteverad.com (0.059); medparkhospital.com (0.044) &                           \textbf{ Shopping }(n = 12, 0.02) &                 vickerypediatrics.com (0.333); babylist.com (0.250); happiestbaby.com (0.250) &                           Education (n = 495, 0.03) &  acog.org (0.255); health.harvard.edu (0.071); sciencedirect.com (0.067 \\
                    News and Media (n = 65, 0.02) &                         nytimes.com (0.215); bbc.co.uk (0.169); today.com (0.123)&         Personal Websites and Blogs (n = 11, 0.02) &                                                                                                                   huckleberrycare.com (1.000) &                           \textbf{Reference (n = 375, 0.03) }&                    quora.com (0.576); en.wikipedia.org (0.171); wikihow.com (0.061) \\
                            Medicine (n = 49, 0.01) &                      goodrx.com (0.755); drugs.com (0.082); yalemedicine.org (0.082)&               Information Technology (n = 6, 0.01) &                                                   drsteverad.com (0.333); utswmed.org (0.333); munchkin.com (0.167)&        Streaming Media and Download (n = 351, 0.02) &                                                                                          youtube.com (0.977); m.youtube.com (0.023) \\
                            \textbf {Shopping} (n = 36, 0.01) &            happiestbaby.com (0.194); babylist.com (0.167); amazon.com (0.139) &                             Medicine (n = 6, 0.01) &                                                                                                  goodrx.com (0.833); yalemedicine.org (0.167) &              Information Technology (n = 256, 0.02) &           utswmed.org (0.477); peanut-app.io (0.141); medparkhospital.com (0.086)\\
        Streaming Media and Download (n = 33, 0.01) &                                                                                                              youtube.com (1.000) &                           \textbf{ Reference (n = 6, 0.01)} &                                                                                                 en.wikipedia.org (0.833); wikihow.com (0.167) &                            \textbf {Shopping} (n = 189, 0.01) &                           amazon.com (0.222); happiestbaby.com (0.180); babylist.com (0.106) \\
                          \textbf{ Reference (n = 32, 0.01)} &          en.wikipedia.org (0.531); quora.com (0.281); pewresearch.org (0.062) &                General Organizations (n = 3, 0.01) &                                                                               center4research.org (0.333); un.org (0.333); unicef.org (0.333) &                      News and Media (n = 187, 0.01) &                        bbc.co.uk (0.118); nytimes.com (0.118); theconversation.com (0.075)\\
         Personal Websites and Blogs (n = 25, 0.01) &                                                                                                      huckleberrycare.com (1.000) &                       News and Media (n = 3, 0.01) &                                                                                                            today.com (0.667); npr.org (0.333) &                            Medicine (n = 157, 0.01) &                     goodrx.com (0.548); drugs.com (0.229); yalemedicine.org (0.083) \\
         \midrule
         \textit{Total} (n = 3779) & - &                      \textit{ Total (n = 490)} &                                                                                                            - &                            \textit{Total} (n = 14634) &                     - \\
\bottomrule
\end{tabular}

\caption{Top Domain Categories and the corresponding top domains of AIO, FS and other ten links.}
\label{tab:top_domain_cateogry}
\end{table*}

\end{document}